**RESEARCH**

**Open Access**

# Supervised topological data analysis for MALDI mass spectrometry imaging applications

Gideon Klaila[1*†], Vladimir Vutov[2†] and Anastasios Stefanou[1†]

†Gideon Klaila, Vladimir Vutov and Anastasios Stefanou have contributed equally to this work

*Correspondence:
klailag@uni-bremen.de

[1] Institute for Algebra, Geometry, Topology and their Applications (ALTA), University of Bremen, 28359 Bremen, Germany
[2] Institute for Statistics, University of Bremen, 28359 Bremen, Germany

**Abstract**

**Background:** Matrix-assisted laser desorption/ionization mass spectrometry imaging (MALDI MSI) displays significant potential for applications in cancer research, especially in tumor typing and subtyping. Lung cancer is the primary cause of tumor-related deaths, where the most lethal entities are adenocarcinoma (ADC) and squamous cell carcinoma (SqCC). Distinguishing between these two common subtypes is crucial for therapy decisions and successful patient management.

**Results:** We propose a new algebraic topological framework, which obtains intrinsic information from MALDI data and transforms it to reflect topological persistence. Our framework offers two main advantages. Firstly, topological persistence aids in distinguishing the signal from noise. Secondly, it compresses the MALDI data, saving storage space and optimizes computational time for subsequent classification tasks. We present an algorithm that efficiently implements our topological framework, relying on a single tuning parameter. Afterwards, logistic regression and random forest classifiers are employed on the extracted persistence features, thereby accomplishing an automated tumor (sub-)typing process. To demonstrate the competitiveness of our proposed framework, we conduct experiments on a real-world MALDI dataset using cross-validation. Furthermore, we showcase the effectiveness of the single denoising parameter by evaluating its performance on synthetic MALDI images with varying levels of noise.

**Conclusion:** Our empirical experiments demonstrate that the proposed algebraic topological framework successfully captures and leverages the intrinsic spectral information from MALDI data, leading to competitive results in classifying lung cancer subtypes. Moreover, the framework's ability to be fine-tuned for denoising highlights its versatility and potential for enhancing data analysis in MALDI applications.

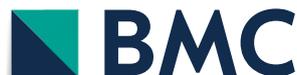

## Background

Matrix-assisted laser desorption/ionization mass spectrometry imaging (MALDI MSI), also known as MALDI Imaging, is a label-free tool for spatially furnishing molecular weight information of compounds like proteins, peptides, and many others (see, e.g., [1, 2]). Provided a thin biological sample (usually a tissue section [3, 4]), MALDI collects mass spectra at multiple discrete positions within the biological sample. As a result, an image is obtained where each spatial spot presents a mass spectrum [5]. The latter depicts the relative abundances of ionizable molecules with a significant number of mass-to-charge ratio (m/z) values, ranging from a couple of hundreds to a few tens of thousands of m/z values (see [3, 5, 6]). MALDI has been demonstrated to be a valuable instrument for many pathological applications (for more details, see [7, 8]), which is possible because of its feasibility in examining formalin-fixed paraffin-embedded (FFPE) tissue samples. In other words, the MALDI tool allows analyses of multiple tumor cores across many patients by aggregating them in a single tissue microarray (TMA) (cf. [2]). As discussed in [9], "the pathological diagnosis of a tumor found in a tissue specimen, including the determination of the tumor origin and genetic subtype" [9] is essential for adequate treatment of patients. This study reports our findings on a MALDI dataset based on two lung cancer (LC) subtypes.

As pointed out by several studies (e.g., in [4, 6, 10]), a plain approach to discovering meaningful m/z values relies upon the idea of identifying significant signal peaks, also known in the literature as peak detection. Focusing on the relevant peaks, one can neglect those highly associated with noise, as acknowledged in [5]. Significant peaks are assumed to provide information for discriminating mass spectra from different cancer subtypes (see, among others, in [11, 12]). Different peak-detection algorithms have been compared in [13]. Furthermore, in [14], a more recent and novel approach proposes to incorporate an isotope pattern [15] around the chosen peaks, which can boost the peak detection methodology.

Other methods aim to extract "characteristic spectral patterns (CSP)" from MALDI-MSI data (cf. [4]). Namely, these methods combine spectral information from different correlated spectral features into a lower dimensional subspace of the data. Afterward, classification models are performed on the extracted feature vectors to classify data units into class labels, i.e., the tumor types or subtypes. Some other frameworks tend to perform feature selection first. Then, based on the selected features, such frameworks execute supervised classification methods to classify observational units into response labels (e.g., [16]). In the context of variable selection, in [10, 17], the authors have proposed approaches by means of large-scale simultaneous testing so as to identify the most associative m/z values with the (cancerous) outcome variables.

The usual challenge modeling methods face is a significant amount of spectral data. As more and more data are being gathered, analyzing the data efficiently with short computational time becomes increasingly more challenging. Topological data analysis (TDA) is a contemporary scientific area that arose from diverse works in applied topology and computational geometry (see [18, 19]). TDA offers an algebraic way of reducing the dimensionality of datasets and extracting essential features in short computational times.

TDA is a relatively novel field of data analysis, and more theoretical work needs to be done to improve the methods (cf. [20]). Even so, TDA has been successfully employed in various fields of science, e.g., in physics, chemistry, and bio-medicine [21], as well as in oncology (see [22, 23]). Motivated by these approaches, this study's objective is to propose a novel framework based on the algebraic topology of MALDI imaging to get improved classification results in a shorter computational time.

In the context of MALDI, one can take advantage of TDA by filtering out the most relevant part of the data, namely the peak-related information. We hypothesize that the importance of a peak increases with its relative height, also known as topological "persistence". Accordingly, low persistent peaks are more likely to be noise. To this end, one can benefit from utilizing our topological framework due to its superior characteristics of denoising and compression.

A general approach to determining a peak's persistence is the upper-level set filtration, where each peak corresponds to a topological feature. These topological features are tracked from their appearance until they merge with a larger feature. This way of analyzing the data is fast in its computational time but has a significant drawback. While tracking the persistence of each peak, one loses the information about their positions, which is paramount to carrying out data analysis applications. For example, in the context of MALDI-related studies, the locations of the biomarkers (cf. [4, 10]) are relevant information for the analysis. To circumvent this limitation, we introduce a different analysis method: the persistence transformation (cf. [24]). The proposed approach keeps track of each peak's position while determining its persistence. This enables the application of this methodology to spectral data.

## Topological data analysis
### MALDI data structure

The first step of our approach is to transform the input MALDI data in order to reflect the topological persistence. MALDI-MSI datasets are commonly stored in an $n \times q$ matrix, denoted by $X$, and $X$ takes its values in $\mathbb{R}_{\geq 0}^{n \times q}$, where each data record corresponds to an intensity value. Usually, mass spectra are stored as rows. While every data column corresponds to an intensity plot for a certain m/z value (see in [9, 25]), $n$ corresponds to the number of mass spectra, and $q$ is the number of m/z values within each mass spectrum. Furthermore, the data records are non-negative since MALDI data presents information on molecular masses of ionizable molecules (cf. [5]). For the convenience of notation, we assume that each mass spectrum is represented as a set of tuples $M := \{(x_1, s_1), (x_2, s_2), \cdots (x_q, s_q)\}$, where $x_j$ is the $j$-th m/z value and $s_j$ is the $j$-th intensity value for $1 \leq j \leq q$. Note that this set $M$ is equipped with a real-valued function $f$ corresponding to the projection to the second coordinate. That means that the intensity values induce a map $f : M \to \mathbb{R}$ with $f(x_j) := s_j$ for $1 \leq j \leq q$ within each mass spectrum.

To sum up, our approach processes each mass spectrum (defined by the set $M$) individually in order to extract the topological properties of the data, i.e., the topological persistence.

**Topological persistence**

Let $f : M \to \mathbb{R}$ be a real-valued function on a compact set, then for $a \in \mathbb{R}$ the upper-level set can be defined as $M_a := \{x \in M | f(x) \geq a\}$. Note that $M_a \subseteq M_{a'}$ for any $a > a'$. This yields the "upper-level set filtration" $M_{a_1} \subseteq \cdots \subseteq M_{a_i}$ for $a_1 > \cdots > a_i \in \mathbb{R}$ (for more details; see Chapter 18 in [20]). In this study, we aim at utilizing the upper-level set filtration instead of the common alternative, i.e., the sublevel set filtration (defined as $M_{\leq a} := \{x \in M | f(x) \leq a\}$). Since the latter, filtration detects local minima and tracks them until they merge with other minima. Conversely, the maxima are highly important in MALDI applications because they correspond to the underlying spectral peaks. As mentioned, peaks provide the necessary information to distinguish mass spectra from different cancerous subtypes [5, 10]. To this end, the upper-level set filtration is of interest in this study since it tracks the local maxima.

Let $x, x' \in M$ with a path-connection, i.e. there exists a continuous function $\rho : [0, 1] \to M$ such that $\rho(0) = x$ and $\rho(1) = x'$. We denote the image of the map $\rho$ by $[x, x']_\rho$. Then we say that $x$ and $x'$ are path-connected in $M_a$, and we write $x \sim_a x'$, if there exists a path connection $\rho$ of $x, x'$, such that $\forall \hat{x} \in [x, x']_\rho : \hat{x} \in M_a$, i.e. $f(\hat{x}) \geq a$, for all $\hat{x} \in [x, x']_\rho$. To track the homology in the upper-level set filtration, we now identify all path-connected points to each other. The degree of the 0-the homology group $h_0$ is given by $b_0(M_a) = |M_a/ \sim_a |$, i.e. the number of different path-connected components in $M_a$. These components are called "topological features". Henceforth, we refer to a (topological) feature in this pure topological sense, not a feature in a statistical sense, like a covariate or an explanatory variable. Remark that there are no features of dimension one or higher since each data unit (mass spectra; see Fig. 5) is represented as a curve.

In the upper-level set filtration, a topological feature is detected for $x$ at $a^*$ if and only if $x \in M_{a^*}$ and $\forall x' \in M_{a^*} : x \not\sim_{a^*} x'$, i.e., $x$ is not path-connected to any other element in $M_{a^*}$. The feature in $x$ merges with another feature in $M_{a^+}$, if $a^+ = \max\{a | x \in M_a \wedge \exists x' \in M_a : x \sim_a x' \wedge f(x') > f(x)\}$, i.e., the largest upper-level set in which the peak gets path-connected to a larger peak. We call $a^*$ the *birth* of the feature, $a^+$ the *death* of the feature, and $p := a^* - a^+$ the *persistence* of the feature. Since the largest peak (the global maximum) does not merge with any other feature, its death is defined as the global minimum. The induced merging order is according to the "elder rule" (see [26]).

To encode all information about features and their persistence in a meaningful and comparable way, the persistence diagram is utilized. Figure 1 displays an example of the encoding process for the upper-level set filtration. Here, the birth and death axis are swapped since the birth values are always greater or equal to the death values. In this way, all the feature points appear above the diagonal line $x = f(x)$ (cf. Section 4 in [27]). In the persistence diagram, each feature is represented by a point $(a^*, a^+)$ with a multiplicity for similar features. The closer a point is to the diagonal of the diagram $\{(x, f(x) = x) \mid \forall x\}$, the lesser its persistence is and vice versa. For more details of the standard persistence diagram approach, we refer to [28, 29].

Given two real-valued functions on a compact set $f, g : M \to \mathbb{R}$, the resulting persistence diagrams **dgm**$(f)$ and **dgm**$(g)$ can be compared with a suitable metric on the diagrams, e.g., the bottleneck distance (cf. [29, 30]). This metric defines the closeness of persistence diagrams and can also indicate the closeness of the corresponding functions

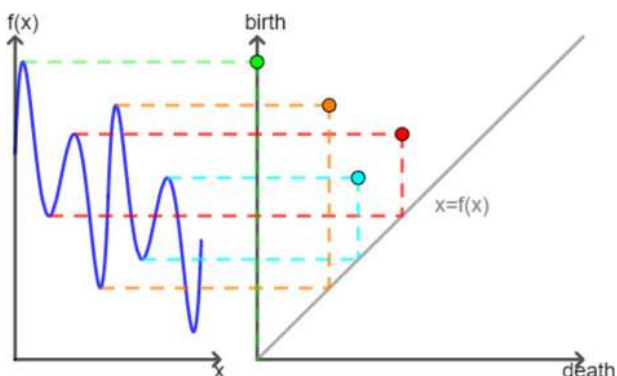

**Fig. 1** The persistence diagram. **Left hand**: A real-valued function $f : M \to \mathbb{R}$ is plotted with $x \in M$ on the x-axis and f(x) on the y-axis. **Right hand**: The corresponding persistence diagram is illustrated. Here, the y-axis represents the birth values, while the x-axis represents the death values. Each point stands for a topological feature for the function f. The first feature, corresponding to the global maxima, never dies. To indicate this, we mark its death value with 0

in the sense that if the persistence diagrams are different, the functions are different. In the context of MALDI-MSI data, the persistence diagram can be used as a proxy for distinguishing cancer (sub-)types. The opposite case that similar persistence diagrams imply similar functions does not hold, as illustrated in Fig. 4.

**Persistence transformation**

Using persistence homology to track the features and applying the bottleneck distance on the resulting persistence diagrams is a viable way to distinguish functions. However, there is a disadvantage to this approach. While tracking the persistence of each feature in a comprehensible way, its position is not being tracked. Nevertheless, as mentioned above, in data-driven applications, including MALDI Imaging, the position of variables is a required property.

To process the information regarding the positions along with the persistence of peaks, we approach the topological manipulation of the spectral data differently. Instead of creating a point $(a^*, a^+) \in \mathbb{R}^2$ for each feature and displaying it in the persistence diagram, we introduce a new dimension to track the position. For each feature, this approach gives a point $(x, a^*, a^+) \in M \times \mathbb{R}^2$, where $x \in M$ is the position of each peak. We define a pairing function $\mu : M \to \mathbb{R}$ with $\mu(x) = a$. The value $a$ is defined to be the highest value smaller or equal to $f(x)$, which upper-level set $M_a$ contains a point $x'$ having a greater function value (similarly as in [24]):

$$\mu(x) := \sup\{a \leq f(x) | \exists x' \in M_a : f(x') > f(x) \wedge x \sim_a x'\}. \qquad (1)$$

For the global maximum $\hat{x}$ the pairing value $\mu(\hat{x})$ is not defined, so we define $\mu(\hat{x}) := \min\{a \in \mathbb{R} | \exists x : f(x) = a\}$ instead. Then the birth of a topological feature is given by $a^* = f(x)$, while the death is calculated by $a^+ = \mu(x)$. The persistence $p$ of a point $x$ can now be defined to be

$$p(x) := f(x) - \mu(x) = a^* - a^+.$$

For any point $x$ not being a local maximum there is a local maximum $x' \in M_{f(x)}$ with $f(x') > f(x)$ and $x \sim_{f(x)} x'$. Then the pairing of $x$ is trivial, i.e. $\mu(x) = f(x)$ with the resulting persistence of $p(x) = f(x) - \mu(x) = 0$. Alternatively, for a point $x$ being a local maximum, the pairing is non-trivial, i.e., there is $\mu(x) < f(x)$, which results in a non-zero persistence $p(x) = f(x) - \mu(x) > 0$. This pairing value always corresponds to $f(\tilde{x})$ for a unique local minimum $\tilde{x}$, i.e.

$$\mu(x) = f(\tilde{x}). \tag{2}$$

The persistence transformation $t : M \to M \times \mathbb{R}^2$ can then be defined for each $x \in M$ to be $t(x) = (x, f(x), \mu(x)) = (x, a^*, a^+) \in M \times \mathbb{R}^2$. For each $x \in M$, the feature triple $t(x) = (x, a^*, a^+)$ consists of the position, the birth value, and the death value. The storage can be reduced to $3 \cdot m$, with $m$ being the number of peaks, by neglecting the trivial tuples, resulting in the "persistence transformation vector".

Similar to the persistence diagram of the upper-level set filtration, the merging of features in the persistence transformation occurs according to the elder rule (see [26]), i.e., the topological feature with the higher birth value persists when merged to another feature. The process can be illustrated in the corresponding merge tree (e.g., in Fig. 2).

**Application**

The notion of the persistence transformation is, in theory, a great way to store more topological information about a graph in general. But many applications do not need the whole information provided by the persistence feature. In these cases, the greater dimensionality of the persistence transformation is rather disadvantageous in calculations. This could be evaded by introducing a "reduced persistence transformation", where instead of $t(x) = (x, a^*, a^+)$ only the position and the persistence are stored:

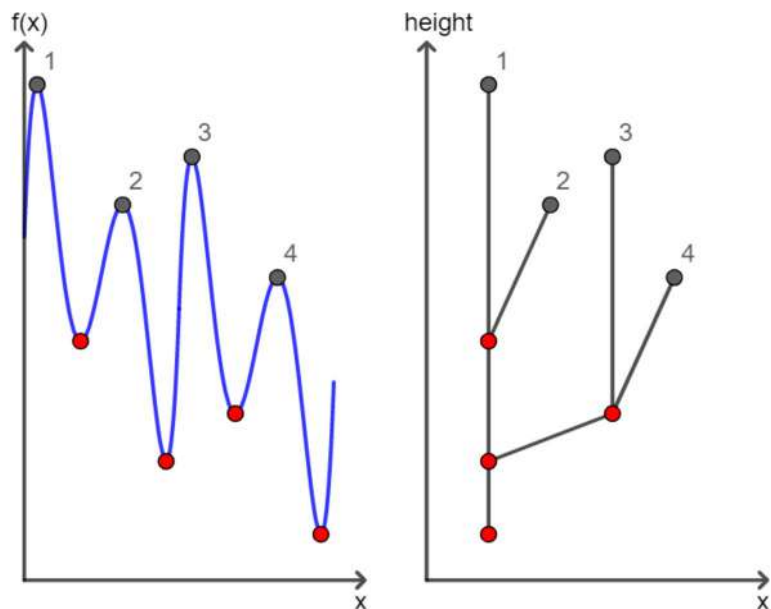

**Fig. 2** The merge tree. A merge tree of a real-valued function $f : M \to \mathbb{R}$ is depicted. On the *x*-axis are the $x \in M$ values. On the *y*-axis is $f(x) \in \mathbb{R}$ (left-hand side) and the height (right-hand side)

$\tilde{t}(x) = (x, a^* - a^+) \in M \times \mathbb{R}$ for $x \in M$. This computational reduction has two benefits. First, it compresses the persistence vector. Second, it can track the persistence of each topological feature and its position.

Another computational improvement for applications can be made by associating low persistent features with noise. By omitting the possible noise by only considering the $k\%$ most persistent features, the accuracy of the analysis can be improved. In addition, reducing the number of stored features might also improve the run-time of subsequent approaches.

Finally, in many applications, there exists a total order on the set $M$, e.g., if $M \subseteq \mathbb{R}$. This order can be passed to the feature space, such that the elder rule for features with equal birth value can be applied deterministically by using the induced order.

**Comparison**

The persistence transformation is strictly a better invariant in distinguishing two functions $f, g : M \to \mathbb{R}$ than the persistence diagram of their 0-dimensional upper-level set filtration. By taking the projection, $p_i((x, a^*, a^+)) = (a^*, a^+)$ of the persistence transformation, the persistence diagram of the upper-level set filtration can be obtained. Hence, all functions that can be distinguished by the persistence diagram can also be distinguished by the persistence transformation. Furthermore, there are cases where the persistence transformation differentiates two functions successfully while the zero-dimensional persistence diagram of the upper-level set filtration fails to do so (see Fig. 3).

The reduced persistence transformation, on the other hand, is not strictly better than the persistence diagram of the upper-level set filtration. However, there are cases where the reduced persistence transformation outperforms the persistence diagram (see Fig. 4). For MALDI-related applications, one seeks to utilize exactly the advantages that the (reduced) transformation offers. For example, the total height of a peak is less interesting in these applications than the relative height, and the position of the peak can be used to backtrack molecules. To this end, in the MALDI-MSI applications, the reduced persistence transformation performs better in analyzing the spectral data than the persistence diagram of the upper-level set filtration.

The benefit of the persistence transformation can be utilized in different scientific areas where the position of the peaks is of interest (e.g., [31]). In the context of TDA, data analyses generally only work reliably if the applied method is stable, meaning that a slight change in the data (given a suitable metric) only leads to small changes in the numerical results. The persistence diagram has been proven stable (see [29]). Furthermore, there are stability theorems for other topological methods (see [32, 33]), but to the best of our knowledge, there has not yet been a proven stability theorem for the persistence transformation. This may be done in further work.

**Implementation and analysis of the Algorithm**

To analyze the MALDI dataset, we implemented a custom-made computer algorithm for the reduced persistence transformation. The recursive algorithm is based on pairing peaks with their unique local minimum (cf. Eq. (2)) to determine their persistence. The pseudo code with a detailed analysis by run-time and storage usage is given in the "Additional file 1", proving the following theorem:

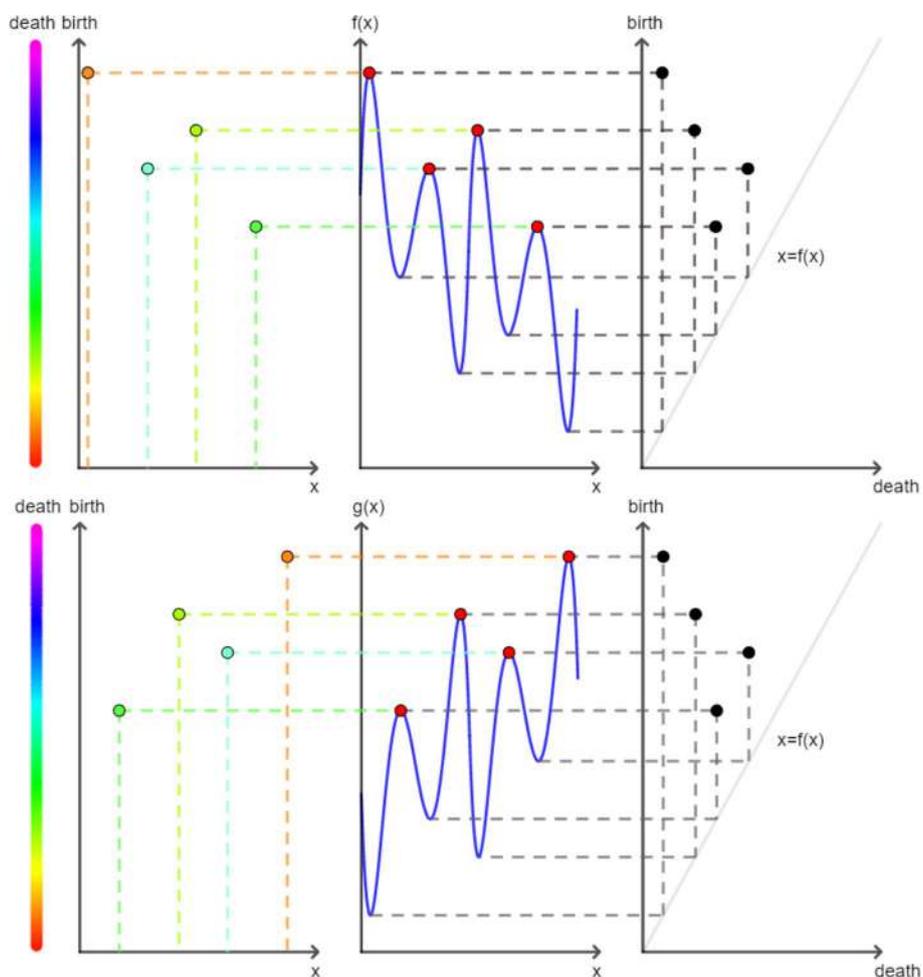

**Fig. 3** The persistence transformation in comparison to the persistence diagram. **Mid**: A real-valued function $f : M \to \mathbb{R}$ (top row) and its mirror equivalent $g : M \to \mathbb{R}$ (bottom row) are illustrated, with $x \in M$ on the x-axis and $y \in \mathbb{R}$ on the y-axis. **Left hand:** The persistence transformation of the functions *f* and *g* are shown with *x* values on the *x*-axis and birth values on the *y*-axis. The color coding of the points corresponds to the third dimension, i.e., the magnitude of death values. The features (i.e., the points) of the functions *f* and *g* can be distinguished clearly. **Right hand**: The persistence diagram of the upper-level set filtration of the functions *f* and *g* is shown with the birth values on the *y*-axis and the death values on the *x*-axis. In contrast to the persistence transformation, the persistence diagrams are identical

**Theorem 1** *The algorithm always terminates and has a complexity of $\sigma(q) + \sigma(m^2)$, where m is the number of peaks for each spectrum and returns all features with their persistence.*

Note that the implementation of the algorithm stores for each feature the tuple $(x, p(x))$, where $p(x) = a^* - a^+$ is the persistence of the feature. Without the further cost of calculation, the algorithm could calculate the persistence transformation instead of the reduced persistence transformation by storing the triple $(x, a^*, a^+)$. Furthermore, the algorithm could be adjusted to determine the persistence diagram of the upper-level set filtration instead by storing the tuple $(a^*, a^+)$.

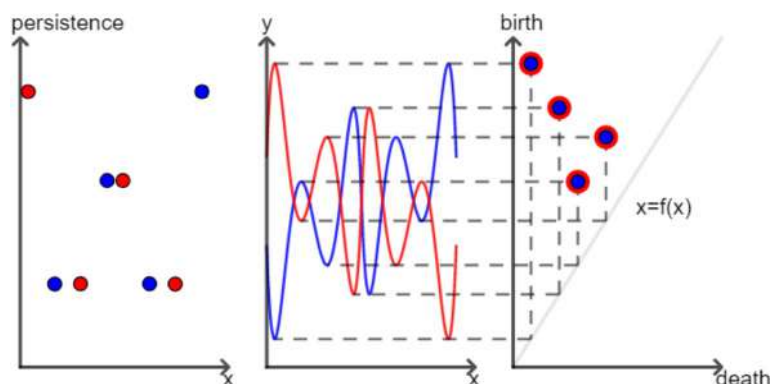

**Fig. 4** The reduced persistence transformation in comparison to the persistence diagram. **Mid**: There are two real-valued functions $f : M \to \mathbb{R}$ (blue) and $g : M \to \mathbb{R}$ (red) plotted with $x \in M$ on the *x*-axis and $y \in \mathbb{R}$ on the *y*-axis. **Left hand**: The persistence transformation of the functions *f* (blue) and *g* (red) are displayed with the *x* values on the *x*-axis and the persistence values on the *y*-axis. The features of *g* are distinct from the features of *f* (blue). **Right hand**: The persistence diagram of the functions *f* (blue) and *g* (red) are shown. The *y*-axis indicates the birth values, while the *x*-axis shows the death values. The features of *g* (red) cannot be distinguished from the features of *f* (blue)

By considering only the peak information while tracking the position of the peaks, the storage space can be compressed to $2 \cdot m$, i.e. the *x* value and the *p* value.

## Supervised methods

The second step of the proposed methodology is to carry out a classification method on the resulting persistences to classify observational units into class labels, i.e., LC subtypes. To do so, we consider two classifiers, logistic regression (LR) and random forest (RF). Note that our goal is to investigate the performance of the proposed topological framework in the context of MALDI modeling, not a benchmark study that compares RF versus LR. For benchmark studies, see, e.g., in [34, 35].

### Logistic regression

Throughout the remainder, we denote the topologically transformed matrix by $Z = (z_{ij})_{\substack{1 \leq i \leq n \\ 1 \leq j \leq q}}$, where the entry $z_{ij}$ corresponds to a persistence value for the *i*-th mass spectrum at the *j*-th m/z value. Let *Y* be a (random) binary outcome variable, meaning that it takes its values in $\{0, 1\}$, in our application, describing two LC subtypes. Further, we denote by $Z_j$ the *j*-th persistence vector corresponding to its *j*-th m/z value alternative. The aim of the logistic regression is to model and estimate the effects of the available covariates on the conditional probability, $\pi_i = P(Y_i = 1 | Z_{i1}, \ldots, Z_{iq})$ for the outcome variable $(Y_i)_{1 \leq i \leq n}$ and the numerical realizations of the covariates $(Z_{ij})_{\substack{1 \leq i \leq n \\ 1 \leq j \leq q}}$. In this setting, the observations $(Y_i, Z_{ij})_{1 \leq j \leq q}$ are assumed to be independent and identically distributed for all $i \in \{1, \ldots, n\}$. LR models combine the probability $\pi_i$ with the linear predictor $\eta_i$ via a "structural" (functional) component given in the linear form $\pi_i = h(\eta_i) = h(\beta_0 + \beta_1 Z_{i1} + \cdots + \beta_q Z_{iq})$ (for more details, see Section 2 in [36]).

In this study, we consider the logit (canonical) link function. Then, the logistic response function is given by

$$\pi_i = h(\eta_i) = \frac{\exp(\eta_i)}{1 + \exp(\eta_i)}.$$

Correspondingly, the logit link function can be expressed as

$$g(\pi_i) = \log\left(\frac{\pi_i}{1 - \pi_i}\right) = \beta_0 + \beta_1 Z_{i1} + \cdots + \beta_q Z_{iq} = Z_i^T \beta. \quad (3)$$

Here, $\beta := (\beta_0, \beta_1, \ldots, \beta_q)^T$ and $Z_i := (1, Z_{i1}, \ldots, Z_{iq})^T$, where the first coordinate corresponds to an intercept term and for all $i \in \{1, \ldots, n\}$. The unknown parameters are (usually) estimated by the maximum (log-) likelihood principle. The log-likelihood function related to model (3) is expressed by

$$l(\beta) = \sum_{i=1}^{n} \{Y_i[\log(\pi_i) - \log(1 - \pi_i)] + \log(1 - \pi_i)\}. \quad (4)$$

For the aforementioned (logit) model, by plugging

$$\pi_i = \frac{\exp(Z_i^T \beta)}{1 + \exp(Z_i^T \beta)} \text{ along with } 1 - \pi_i = \frac{1}{1 + \exp(Z_i^T \beta)}$$

in (4), it yields that

$$l(\beta) = \sum_{i=1}^{n} \{Y_i\left(Z_i^T \beta\right) - \log\left(1 + \exp(Z_i^T \beta)\right)\}. \quad (5)$$

Finally, the probability that an unobserved (new) observation is assigned to class 1 is estimated by substituting $(\beta_0, \ldots, \beta_q)$ by their fitted counterparts and $Z$'s by their numerical realizations for the considered new observation in the conditional $P(Y = 1|Z_0, \ldots, Z_q) = \frac{\exp(Z^T \beta)}{1+\exp(Z^T \beta)}$, where we have $q + 1$ covariates since the first coordinate corresponds to the intercept term. Respectively, the new observation is assigned to class $Y = 1$ if the conditional probability, $P(Y = 1) > c$, is greater than a pre-specified threshold $c$, and oppositely to class $Y = 0$. In this study, we set $c = 0.5$ – a commonly used threshold (cf. [34]). To obtain the numerical results, we adopted the "LogisticRegression()" function in *scikit-learn, v. 1.2.1* [37] (with no penalty) to obtain our numerical results.

**Random forest**

The RF algorithm has become an established non-parametric procedure for regression and classification tasks. It has been broadly used in various scientific disciplines [38–40], including subtyping of lung cancer [41]. RF was originally introduced by Leo Breiman in [42] and it presents an "ensemble learning" approach constituting the aggregation of a collection of a great number of decision trees [43]. RF takes advantage of numerous decision trees, which leads to a reduction of empirical variance in comparison to a single (decision) tree and significant enhancements in its prediction accuracy [35]. RF utilizes decision trees in order to calculate the majority votes in the leaf nodes when deciding a class label for each observational unit [35, 42]. In essence, RF consists of two steps. The first step is to build an RF tree. The following step is to classify the data on the basis of an RF tree that has been generated in the first step. For more details, e.g., see in [39, 44].

In this study, we employ the original variant of RF (see [42]), where each tree of the RF algorithm is constructed on a bootstrap sample drawn arbitrarily from the data by employing the classification and regression trees method and minimizes the Gini impurity (GI) regarding the splitting criterion. When constructing each tree (for each split), solely a pre-specified number of randomly selected (data) covariates are deemed as candidates for splitting.

An important step when using RF is selecting hyperparameters, also called tuning parameters. Their values have to be optimized attentively since the optimal quantities depend on the data at hand. An essential concept regarding tuning optimization is "overfitting". In other words, tuning parameters related to complex rules be inclined to "overfit" the training data. As a result, they produce prediction rules overly specific to the training data, performing well for that (training) data but potentially underperforming when applied to independent data. As discussed in [45], the choice of less-than-optimal parameter quantities can be (at least) partially prevented by utilizing a test set or cross-validation (CV) procedures for tuning. However, it is out of the scope of this study to identify the (most) optimal tuning parameters in the context of MALDI modeling. Instead, we are predominantly interested in evaluating the performance of the proposed TDA approach. To this end, we select the "typical default values" for the RF algorithm, as listed in Table 1 in [45]. Specifically, we set the tuning parameters in our numerical experiments as follows: the number of trees equals 1000. The number of drawn candidate variables per split is equal to $\sqrt{q}$ (often referred to as "mtry", *max_features* in *scikit-learn*). The splitting criterion in the nodes is the Gini impurity. The minimum number of samples in a terminal node is equal to one (*min_samples_leaf* in *scikit-learn*). Regarding the sampling scheme, the number of observational units that are (randomly) drawn for training each tree is determined by the sample size parameter. The default value corresponds to *n* (i.e., to the overall number of data samples), respectively, observational units are drawn with replacement when generating each tree. The seed for all experiments was set to 1234.

We carried out the RF approach on the basis of the resulting algebraic vectors $Z := (Z_{i1}, \ldots, Z_{iq})^T$ and the binary outcome $Y_i$ for all $i \in \{1, \ldots, n\}$. Notice that a unit vector has not been considered, i.e., without an intercept term, as in the case of LR. We adopted the function '*RandomForestClassifier()*' in [37] (*version 1.2.1*) to yield the numerical results.

## Real data analysis
### Description of the MALDI-MSI data

Here, we present the empirical results obtained by applying multiple classification schemes to MALDI-MSI data based on different levels of persistence extraction. Several studies have previously analyzed this dataset (cf. [4, 6, 7, 9, 10]). We refer to [4, 7, 9] for an in-depth description of the aforementioned dataset concerning its acquisition protocols, tissue sections, tissue blocks, etc. Here, we provide only a brief outline of the dataset.

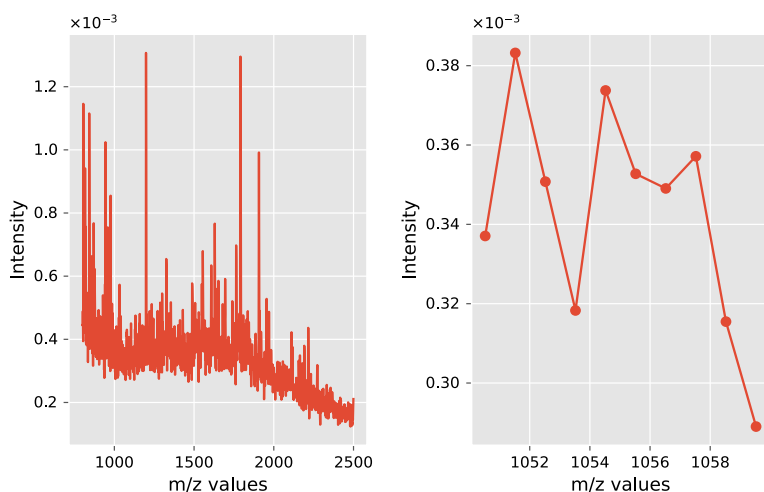

**Fig. 5** An example of a mass spectrum. **Left hand:** An example of a mass spectrum from a single cancerous spot within a patient tissue. **Right hand:** A closer look at spectral data following spectral filtering

**Table 1** Descriptive statistics for each TMA

| TMA | Number of spectra | Ratio ADC: SqCC (%) |
| --- | --- | --- |
| $TMA_1$ | 680 | 59.3 |
| $TMA_2$ | 437 | 60.4 |
| $TMA_3$ | 563 | 41.2 |
| $TMA_4$ | 601 | 49.3 |
| $TMA_5$ | 512 | 63.5 |
| $TMA_6$ | 650 | 76.8 |
| $TMA_7$ | 536 | 50.4 |
| $TMA_8$ | 690 | 54.3 |

Cylindrical tissue cores (CTCs) of non-small cell lung cancer were taken from 304 patients. Specifically, 168 patients were associated with primary lung adenocarcinoma (ADC), while 136 patients were associated with primary squamous cell carcinoma (SqCC). CTCs of all patients were gathered into eight tissue microarray (TMA) blocks (for descriptive statistics, see Table 1). As discussed in [9], the tumor status and subtyping for all CTCs were affirmed by standard "histopathological examination". Furthermore, this dataset has been generated only on annotated subregions called regions-of-interest (cf. [6]), i.e., subregions comprising only tumor cells.

For illustrative purposes, Fig. 5 depicts an example of an output from a MALDI experiment taken from a single spatial location in the provided tissue. In the left panel of Fig. 5, the m/z values are illustrated on the x-axis, while the intensity values of "ionizable molecules" are charted on the vertical axis. This spatial information can be used in two directions, namely, for the determination of the subtyping (i.e., the cancer subtypes) or the identification of the source of the tumor in tissue. The right panel of Fig. 5 illustrates the data granularity following one of the data-processing steps, i.e., spectral filtering. Namely, the latter means that m/z values were centered around their expected peptide masses (for more details, see [9] and the references therein). Other data-processing steps applied to this dataset were baseline correction and total ion count (TIC) normalization.

As pointed out in different studies (e.g., [46, 47]), LC is the primary cause of cancer-related fatalities globally; for example, there were 1.59 million reported deaths in 2012 (see [47]). Two major LC categories are recognized, i.e., small cell lung cancer (SCLC) and non-small cell lung cancer (NSLC). The latter constitutes approximately 85% of all LC cases as reported. The two prevailing NSLC entities are ADC and SqCC, compromising approx. 50% and 40% of all lung-related cancers, respectively [46, 47]. As discussed in [9], the distinction between these two common subtypes is of great importance for the therapy choice of patients.

The used dataset can be found on Gitlab, as provided in [9]. Note we did not apply any other data-processing steps on this dataset. Specifically, this dataset contains $n = 4669$ (observational units, the number of mass spectra), and the number of m/z values is $q = 1699$.

**Classification evaluation**

To evaluate the performance of the classification schemes, we mimic the realistic scenario proposed in [9]. Namely, the data is split into training and test sets. The upcoming results were derived by performing k-fold cross-validation (CV) on a TMA level. We followed both scenarios as proposed in [9], specifically 8-fold CV and 2-fold CV. Regarding the 8-fold CV, eight distinct test subsets were created based on each TMA from the overall set of eight TMA blocks. Then, in each of the eight CV folds, each classification scheme was applied to seven TMAs and predicted on the remaining test set—not considered in the training process. Likewise, we carried out 2-fold CV, creating two subsets $A := \{TMA_1, \ldots, TMA_4\}$ and $B := \{TMA_5, \ldots, TMA_8\}$. We reported the obtained results on the basis of all test sets for the 2-fold and 8-fold CVs.

The classification accuracy was assessed by computing the balanced accuracy. The latter metric is computed as the average proportions of correctly classified spectra for each class separately. As a result, this metric is independent with respect to imbalanced binary categories, i.e., when one of the target classes appears far more often than the other in the test set. To illustrate the numerical performances appertaining to the persistence transformation, we set a tuning parameter $k$ based on different percentages of peaks extraction.

**Data-analysis results**

Figure 6 depicts an example of the proposed topological feature extraction. Namely, the top row illustrates an example of a raw spectrum (cf. Figure 5), whereas the next subplots depict extracted persistence values. Apart from the first row, the m/z values are plotted on the horizontal axes, while the vertical axes show the derived persistence values, not the intensity values as in the first row of the figure.

Tables 2, 3 summarize the classification results obtained by applying LR and RF classifiers based on several levels of the extracted persistence vectors. The percentage of extracted persistence employed for the evaluation, denoted by $k$, ranges in a grid of pre-specified percentages—$k \in \{10\%, 20\%, 25\%, 30\%, 40\%, 50\%\}$. By compressing the raw

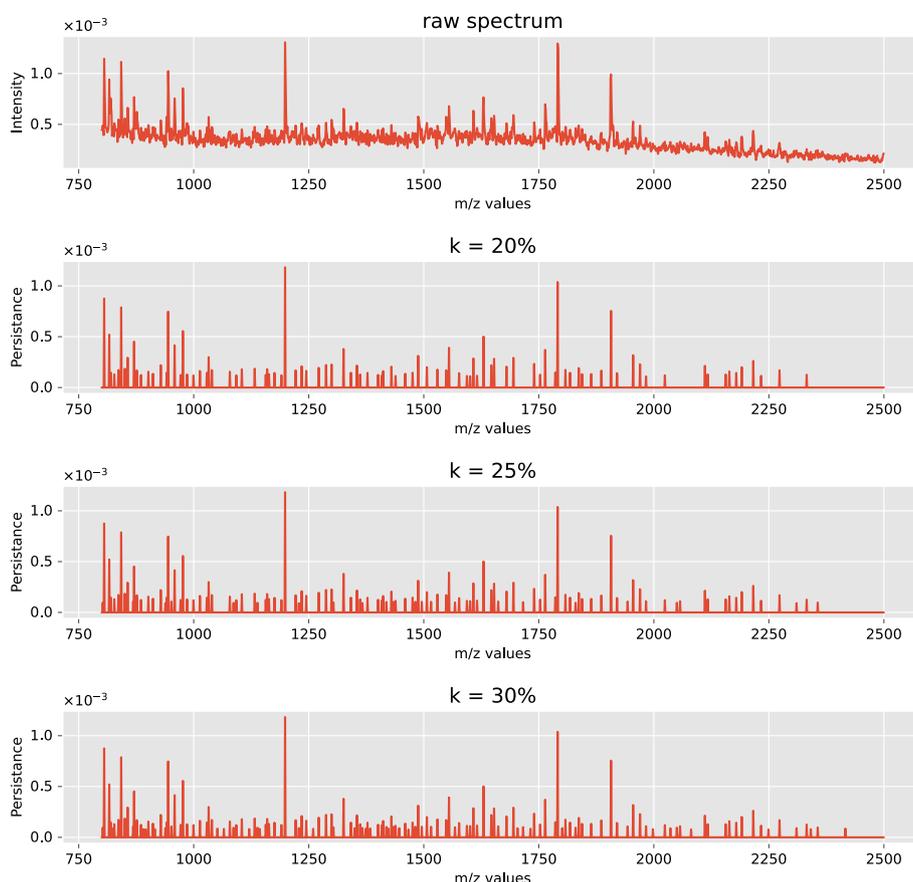

**Fig. 6** Application of the reduced persistence transformation. A visualizing inspection of the proposed topological framework. The top row subplot depicts the raw spectrum, while the following subplots illustrate simplified representations of this raw spectrum given different percentages of peak extraction (abbreviated with *k*%). Namely, the raw spectrum is transformed to sparse tuples based on a pre-specified k level of peaks extraction

**Table 2** Comparison table of 8-fold cross-validation. The table tabulates the obtained balanced accuracy results given different percentages of peaks extraction and non-extracted data based on the LR classifier

| Statistic | k = 10% | k = 20% | k = 25% | k = 30% | k = 40% | k = 50% |
|---|---|---|---|---|---|---|
| mean | 0.826 | 0.845 | 0.859 | 0.866 | 0.859 | 0.859 |
| min | 0.722 | 0.716 | 0.724 | 0.743 | 0.737 | 0.723 |
| max | 0.917 | 0.937 | 0.931 | 0.950 | 0.935 | 0.935 |
| median | 0.827 | 0.869 | 0.886 | 0.888 | 0.880 | 0.878 |
| std | 0.074 | 0.079 | 0.077 | 0.076 | 0.076 | 0.078 |

data with respect to different *k*, we observed a significant gain in the computational time for executing RF. For example, to execute the 8-fold CV task on 7 CPUs on a standard machine, it takes 35 seconds when using $k = 5\%$ and 70 seconds when using $k = 50\%$, in contrary to the 215 seconds it takes using the raw data, i.e., the original variant where each data entry corresponds to intensity value.

Table 3 Comparison table of 8-fold cross-validation. The table tabulates the obtained balanced accuracy results given different percentages of peaks extraction and non-extracted data based on the RF classifier

| Statistic | k = 10% | k = 20% | k = 25% | k = 30% | k = 40% | k = 50% |
|---|---|---|---|---|---|---|
| mean | 0.831 | 0.863 | 0.871 | 0.878 | 0.877 | 0.868 |
| min | 0.702 | 0.734 | 0.760 | 0.774 | 0.776 | 0.746 |
| max | 0.936 | 0.945 | 0.940 | 0.959 | 0.930 | 0.949 |
| median | 0.849 | 0.873 | 0.880 | 0.892 | 0.900 | 0.890 |
| std | 0.083 | 0.076 | 0.066 | 0.067 | 0.061 | 0.079 |

To put our results in the context of other competitors for this dataset, we performed a comparison with a popular method for retrieving informative parts of the spectral data and executing automated cancer (sub-)typing. Briefly put, in [9], the authors proposed novel supervised non-negative matrix factorization (NMF) methods: the classification tasks are executed in parallel to feature extractions (in the context of NMF extraction), which differs from the more classical NMF-related scenario (cf. [4]). The authors introduced 13 distinct classification schemes. From these, we selected the top 2 competitors; for the remaining ones, we refer interested readers to Figure 3 and Figure 4 in [9]. These top 2 competitors from [9] are: *Flog_int Flog_log*, where the number of NMF "features" is 60, as suggested in the provided code. Our competitors are the persistence transformation where $k$ is either 30% or 40% and the FR classifier. These schemes are abbreviated to *PT_RF*_30% and *PT_RF*_40%, respectively.

Table 4 illustrates that *Flog_int* and *Flog_log* perform slightly better than our best performers for this dataset for the 8-fold CV task and 2-fold Train B. However, the topological-based competitors outperformed the other 11 NMF-based schemes for this dataset, even in some scenarios *PT_RF*_40% produces numerically similar results vis-à-vis all NMF-based competitors, cf. Bal. Acc. (2-fold) Train A. The authors [9] concluded that apart from the top three classification schemes, most of the other methods achieved, on average, balanced accuracy values below 80%. This table stands for the proof of concept that our topological framework accompanying the RF classifier can produce competitive results. Moreover, the RF (non-linear) algorithm can operate in pure high-dimensional scenarios, i.e., when covariates exceed the observational units $n \ll q$. Therefore, the proposed classification scheme $PT + RF$ can also be applied in different data regimes.

## Image denoising with persistence transformation
### Simulation setup

A (big) challenge in examining real-world applications is the presence of noise that can corrupt the data and lead to incorrect data-analysis results (see [14, 48]). To this end, researchers have to be careful when analyzing datasets with a possibility of noise and address it appropriately to improve the accuracy of the results (see [10, 17, 49]).

To assess the effectiveness of the proposed topological framework given the presence of noise, we proceeded as follows. First, we simulated multiple synthetic mass spectrometry (MS) images, where each pixel of these images corresponds to a unique mass spectrum. Specifically, each pixel presents an average value for its respective mass spectrum.

**Table 4** Performance of the proposed classification algorithms vis-à-vis the best classification competitors from [9]. *PT_ RF_k*% stands for Persistence Transformation, using the Random Forest classifierer with *k* as hyperparameter. According to [9], Flog stands for the Frobenius norm utilizing the logistic regression classifier. The suffix '*_int*' denotes the Integrated approach, while '*_log*' stands for the Optimized approach

| Method | Avg. Bal. Acc. (8-Fold) | Bal. Acc. (2-Fold) Train A (%) | Bal. Acc. (2-Fold) Train B (%) |
|---|---|---|---|
| *PT_ RF_*30% | 87.8% ± 6.70 | 90.75 | 84.40 |
| *PT_ RF_*40% | 87.7% ± 6.01 | 91.15 | 83.53 |
| *Flog_int* | 89.8% ± 4.35 | 90.8 | 89.1 |
| *Flog_log* | 88.8% ± 6.36 | 91.1 | 87.1 |

Second, we artificially contaminated the spectral data that generated the MS images by adding different types and levels of noise. Finally, for each figure, we plotted the ground truth, the noise image, and two variants of denoised MS images in a row. As a result, one can pictorially identify the ability of our TDA approach to differentiate signal from noise in the images. Two of these results are displayed in Fig. 7 and in Fig. 9.

We utilized the *"Cardinal"* package ([50] *v. 3.0.1*) in *R* ([51] to simulate noiseless (ground truth) MS images. Accordingly, we employed the "SimulateImage()" function with the following parameters: the preset image is two (i.e., there are two figures, a circle in the top-left corner and a square in the bottom-right corner), the m/z range lies in $500 - 2000$ (resulting in 3466 m/z values), the number of peaks $k^* = 50$, and a noiseless image (i.e., "sdnoise" equals to zero). We aim to demonstrate the efficacy of our methodology with varying baseline levels so as to cover baseline value ranges $\in \{0, 5, 15\}$. Based on these parameters, we simulated multiple MS images with sizes $\{30 \times 30, 42 \times 42, 60 \times 60\}$. Following the simulation of the ground truth images, we contaminated the spectral data by adding either Gaussian or Poisson noise. We chose increasing values for the standard deviation for the Gaussian noise and $\lambda$ for the Poisson noise, i.e., artificially contaminated synthetic data more and more. These distribution parameters are given in the caption of each subplot and can serve as a proxy for different signal-to-noise ratio variants. Due to its relevance to our real-world data application, we proceeded by picking a percentage of the most significant peaks – signals outside these fractions were considered noise.

Adding artificial noise to the MALDI spectra has two significant effects. First, the height of the existing signal peaks could be altered. Second, new noise peaks can be established. For low levels of noise, the added noise peaks are less persistent than the signal peaks. By optimizing the tuning parameter *k*, our algorithm can differentiate the bulk of the signal peaks by neglecting the noisy ones in spectral data. A good example of low-level noise is the Gaussian noise, which is displayed in Figs. 7 and 8. As can be seen in Fig. 8, most of the signal peaks can be distinguished from the noise peaks by their height. Hence, the original shapes from the ground truth images can be reconstructed in the denoised images in Fig. 7 when the tuning parameter *k* is chosen accordingly.

Higher levels of noise, on the other hand, can be challenging for the persistence transformation. By adding noise peaks larger than the signal peaks, the ground truth can be compromised so that the persistence transformation can not reconstruct the original

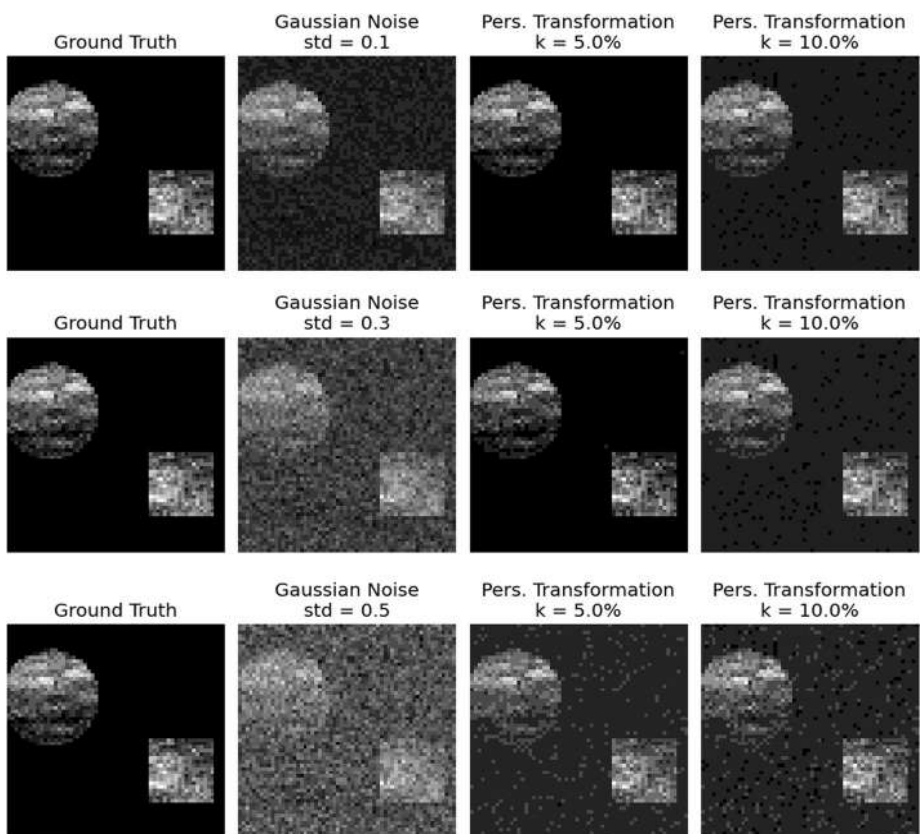

**Fig. 7** Denoising with the persistence transformation. On the leftmost column, the ground truths of synthetic MALDI-Images are displayed. In the second column, distinct levels of Gaussian noise are added to the spectral data. The processed images based on two choices of *k* are displayed in the third and fourth columns

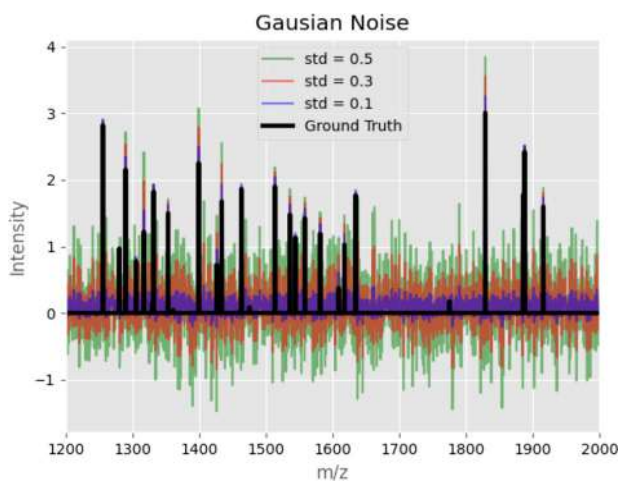

**Fig. 8** An example of a synthetic spectrum of the images in Fig. 7. The ground truth spectrum is displayed in black, and its different noisy counterparts are displayed in color

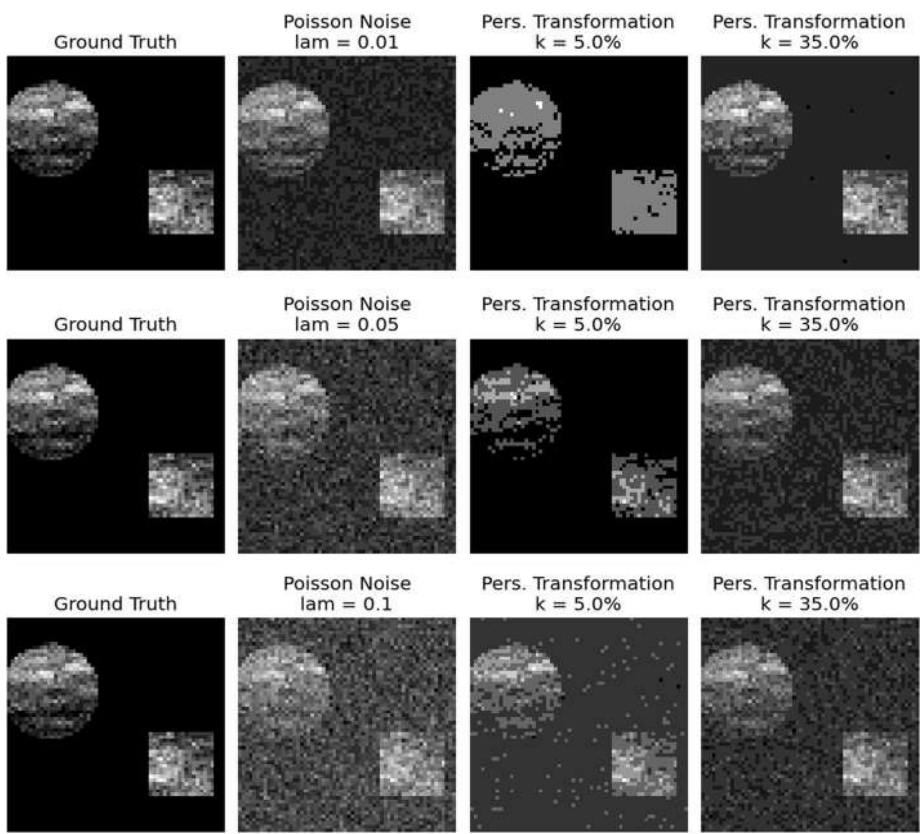

**Fig. 9** Denoising with the persistence transformation. On the leftmost column, the ground truths of synthetic MALDI-Images are displayed. In the second column, distinct levels of Poisson noise are added to the spectral data. The processed images based on two choices of *k* are displayed in the third and fourth columns

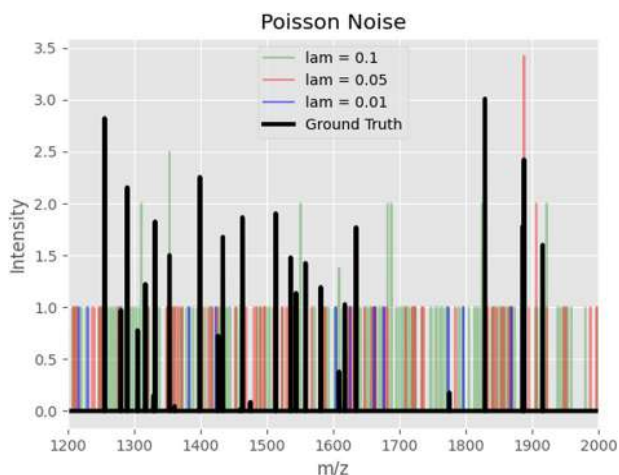

**Fig. 10** An example of a synthetic spectrum of the images in Fig. 9. The ground truth spectrum is displayed in black, and its different noisy counterparts are displayed in color

shapes. An example of such kind of noise is the Poisson noise, and it is displayed in Figs. 9 and 10. In Fig. 10, it is shown that most of the noise spectra are exceeding the signal peaks. Even so, with an appropriate *k* value, the shapes of the ground truth can be reconstructed to some degree in Fig. 9.

More results can be found in the "Additional file 2". We want to highlight the time used for the (denoising) analysis. It can be seen that doubling the number of pixels results in doubled time for the algorithm, e.g., for Gaussian noise with a standard deviation of 0.1, the analysis of a $30 \times 30$ image takes approx. 29 seconds, for a $42 \times 42$ image it takes approx. 61 seconds, and for a $60 \times 60$ image it takes approx. 113 seconds on a standard computer. This illustrates the almost linearity of the implementation.

## Discussion

### Summary

Motivated by the MALDI classification studies, the objective of this study has been to propose a novel custom-made approach for modeling MALDI-MSI data. In general, the study's methodology consists of two steps. First, we carry out the introduced topological framework to obtain the intrinsic information from each mass spectrum, given thousands of m/z values. Generally speaking, this step can be considered as a data-compression method for MALDI-MSI data. Second, we execute two supervised classification methods based on the resulting persistence vectors so as to classify the observational units into lung cancer subtypes.

The usefulness of the proposed topological framework consists of three perspectives. First, our numerical classification results illustrate that the topological framework extracts the necessary information, which can be used for further classification tasks. The obtained results are competitive with other data-analysis methods for this dataset (cf. [9]). Second, the proposed framework compresses MALDI-MSI data, resulting in a significant computational gain for the RF classifier. Third, we have demonstrated its effectiveness in retrieving the informative parts of spectral signals under different noisy scenarios. The proposed topological framework can be adopted in a computationally efficient algorithm depending on a single tuning parameter, i.e., the fraction of used peaks.

### Outlook

The persistence transformation is a novel tool for topological data analysis, which can be employed in different real-world applications. Future work to extend the introduced framework might include the application of the heat equation to reduce the impact of noisy peaks (see [24]), which might increase the classification performance. Alternatively, one can use the persistence pairs as input points for a second persistence homology analysis. However, any computational improvement would lead to the extension of the computational time. Furthermore, the information on the position of the peaks might be used for backtracking to identify (biologically) relevant molecules given as peaks. Finally, the algorithm used in this paper shows similarity to Morse Theory [52] and especially to the matching theorems in [20]. This presents an interesting topic for follow-up research.

## Supplementary Information

The online version contains supplementary material available at https://doi.org/10.1186/s12859-023-05402-0.

**Additional file 1**. The file contains the pseudo-code of the introduced algorithm and the proof of its complexity and correctness.

**Additional file 2**. The file contains synthetic MALDI-images. Distinct types and levels of noise are added to the ground truth and displayed. Finally, the results of the denoising with the persistence transformation is depicted.

### Acknowledgements
We thank the editor-in-chief, the associate editor, and two anonymous reviewers for their reading of the paper and for their constructive suggestions, which have improved the presentation. The authors extend their sincere gratitude to Prof. Dmitry Feichtner-Kozlov and Prof. Thorsten Dickhaus at the University of Bremen for supervising GK and VV. Also, we sincerely thank Lena Ranke, Friederike Preusse, and Lukas Mentz for their constructive criticism of the manuscript. The financial support by the German Research Foundation via the RTG 2224, titled "$\pi^3$: Parameter Identification—Analysis, Algorithms, Implementations" is gratefully acknowledged.

### Author contributions
This paper is based on joined work of the three main authors as equal contributors. GK implemented the algorithm and proved the complexity. VV took care of the statistical treatment. AS helped clarify the notion of persistent transformation and its connection to the persistence diagram via the elder rule and proposed examples illustrating this connection. All authors have written the manuscript and approved the final version.

### Funding
Open Access funding enabled and organized by Projekt DEAL. Deutsche Forschungsgemeinschaft. Grant Number: RTG 2224

### Availability of data and materials
Python code with the numerical results presented in the paper, including the MALDI data, is available at: https://github.com/klailag/SupervisedTDAMethodForMALDI

## Declarations

**Ethics approval and consent to participate**
Not applicable.

**Consent for publication**
Not applicable.

**Competing interests**
The authors declare that they have no competing interests.

Received: 24 February 2023   Accepted: 26 June 2023
Published online: 10 July 2023

# Additional file 1 — The introduced algorithm

This additional file provides the pseudo-code of the algorithm introduced in the paper *Supervised topological data analysis for MALDI mass spectrometry imaging applications.* Furthermore, it contains the proof of Theorem 1 of the paper.

---

**Algorithm 1** Recursion start

1: **Input:** $[f(x_0), \ldots, f(x_{q-1})]$
2: **Return:** $[(\hat{x}, p(\hat{x})), \ldots]$
3: maxima, minima, featurePairs $\leftarrow \varnothing$
4: **for all** $x_j \in [x_0, x_{q-1}]$ **do**
5:    **if** $x_j$ is maximum **then**
6:       maxima $\leftarrow (x_j, f(x_j))$
7:    **else if** $x_j$ is minimum **then**
8:       minima $\leftarrow (x_j, f(x_j))$
9: SORT(maxima, f(x), >)
10: SORT(minima, f(x), <)
11: $(\hat{x}, f(\hat{x})) \leftarrow$ maxima.pop(0)
12: featurePairs $\leftarrow (\hat{x}, f(\hat{x}) -$ minima[0][1])
13: RecursionStep($x_0, \hat{x}$, maxima.copy(), minima.copy(), featurePairs)
14: RecursionStep($x_n, \hat{x}$, maxima.copy(), minima.copy(), featurePairs)
15: **return** featurePairs

---

**Algorithm 2** Recursion Step

1: **Input:** start, end, maxima, minima, featurePairs
2: **for all** $(x_j, f(x_j)) \in$ maxima **do**
3:    **if** $x_j \notin [\text{start}, \text{end})$ **then**
4:       maxima $\leftarrow$ maxima$\setminus(x_j, f(x_j))$
5: **if** $|\text{maxima}| = 0$ **then**
6:    **return**
7: $(\hat{x}, f(\hat{x})) \leftarrow$ maxima.pop(0)
8: RecursionStep(start, $\hat{x}$, maxima.copy(), minima.copy(), featurePairs)
9: **for all** $(x_j, f(x_j)) \in$ minima **do**
10:    **if** $x_j \notin (\hat{x}, \text{end})$ **then**
11:       minima $\leftarrow$ minima$\setminus(x_j, f(x_j))$
12: $(x', f(x')) \leftarrow$ minima.pop(0)
13: featurePairs $\leftarrow (\hat{x}, f(\hat{x}) - f(x'))$
14: RecursionStep($x', \hat{x}$, maxima.copy(), minima.copy(), featurePairs)
15: RecursionStep($x'$, end, maxima.copy(), minima.copy(), featurePairs)

---

*Proof of Theorem 1:* The algorithm to calculate the reduced persistence transformation is divided into two parts: the recursion start (Algorithm 1) and the recursion step (Algorithm 2).

The *recursion start* (Algorithm 1) gets as input the list of all the intensity values for each m/z value, marked as $f(x_j)$. Let $m$ be the number of maxima in this mass spectrum. In the beginning, empty lists are created for the maxima, the minima, and the results (called "featurePairs"). The latter list stores the $x$ value, i.e., the position, as well as the persistence of each peak. Notice that each recursion step updates the list instead of returning results.

In the next step, all the minima and the maxima are stored in the corresponding lists in tuples of the form $(x, f(x))$. For this, the algorithm iterates through the list of the $f(x_j)$. If the value is larger than its neighbors, it is marked as maximum and stored in the corresponding list. Correspondingly, values that are smaller than their neighbors are marked as a minimum. This identification of extremal points can be made in linear run-time since the list is traversed just once, resulting in a complexity of $\sigma(q)$. The two lists *maxima* and *minima* are then sorted by their corresponding value $f(x_j)$ (the *minima* list inverted) with a complexity of $\sigma(m \cdot \log m)$.

For the largest feature, the global maximum, the persistence is defined to be the difference to the global minimum (see Equation (1) of the paper *Supervised topological data analysis for MALDI mass spectrometry imaging applications*). These values are the first elements of their corresponding lists. After calculating the persistence, the maximum is removed from the list, and the found feature $(x, p(x))$ is stored in the list *persistencePairs*. The *recursion step* is called afterwards with the intervals $[x_0, \hat{x})$ and $[x_{q-1}, \hat{x})$. Notice that the second interval is reversed. As input, the *recursion step* gets a copy of the two lists minima and maxima as well as the original list *featurePairs*. All these computations can be done in constant time, i.e., $\sigma(1)$. After the last *recursion step*, all the features are detected and stored in the *featurePairs* list and can be returned.

The input for the *recursion step* (Algorithm 2) consists of two indices, namely *start* and *end* (indicating the part of the data which is processed in the current recursion step, i.e., the positions of m/z values), a list of *maxima* and a list of *minima*, and the shared list of *featurePairs*. In the first step, the routine removes all the maxima not in the currently processed part of the data. There are at most $m$ elements in the *maximum* list, so the complexity of this task is $\sigma(m)$. If the list is empty after the removing step, the *recursion step* reaches the end and can return. If not, it removes the first element $\hat{x}$ from the list. This is the most persistent feature (in terms of topology) in the processed part of the data. The elder rule (cf. [1]) states that the feature can only merge with a feature with a larger persistence, which is per construction at the index *end*. The corresponding minimum (cf. Equation (2) of the paper *Supervised topological data analysis for MALDI mass spectrometry imaging applications*) to $\hat{x}$ can only be in the interval $(\hat{x}, end)$ so the *minima* list can be filtered in a similar fashion to the *maxima* list with the same complexity. Since there are more possible features in the interval $(start, \hat{x})$, the *recursion step* is called once more for this interval.

The values $\hat{x}$ and the smallest minimum $x'$ from the *minima* list generate a topological feature. This feature is updated in the original *featurePairs* list, and the recursion step can be repeated with the two intervals $(x', \hat{x})$ and $(x', end)$.

The recursion step is processed at least once for each maximum with three extra calls after no more maxima are left, i.e., it runs at most $4 \times m$ times. Given the complexity of each step of $\sigma(m)$, the complexity of all the recursion steps together is $\sigma(m^2)$. This gives an overall complexity of the algorithm of

$$\sigma(q) + \sigma(m^2) + \sigma(m \cdot \log m) = \sigma(q) + \sigma(m^2).$$

For each maximum, there is a tuple stored which contains the information of the position and the persistence, resulting in an overall storage use of $2m$ elements.

The algorithm always terminates since, at each recursion step, one maximum is removed from the list of maxima —if it is not already empty. Likewise, each part of the input list is being processed. Since there is only a finite number of elements in the *maxima* list (i.e., $m$), the algorithm terminates after all are processed. Even more, the algorithm returns all the features with their persistence. Each maximum creates a feature, and all maxima are processed. They are paired with the correct minimum between themselves and a feature with a higher persistence according to the elder rule (see [1]). Hence, the algorithm always terminates and returns the correct solution in $\sigma(q) + \sigma(m^2)$ run-time. □

# Additional file 2 — Additional simulation results

The additional file contains synthetic MALDI-images. Distinct types and levels of noise are added to the ground truth and displayed. Finally, the results of the denoising with the persistence transformation is depicted.

**Gaussian Noise**

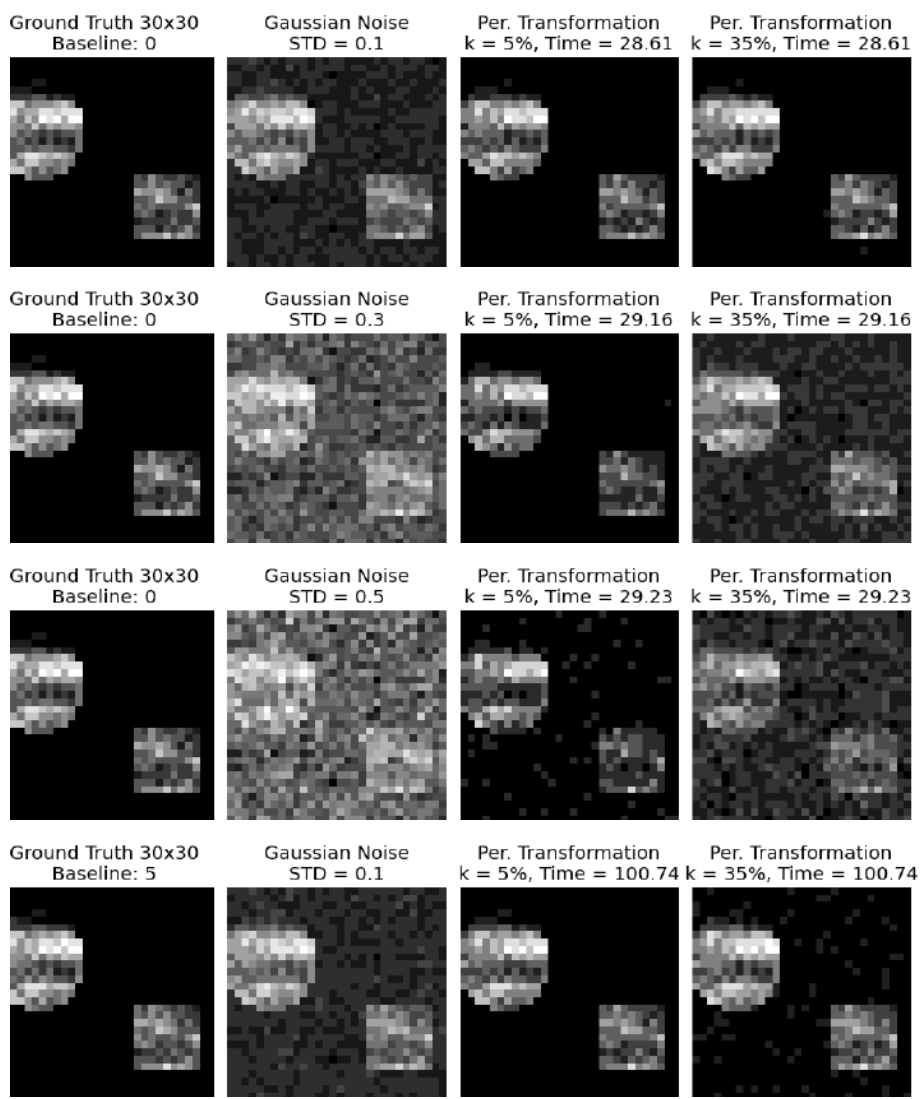

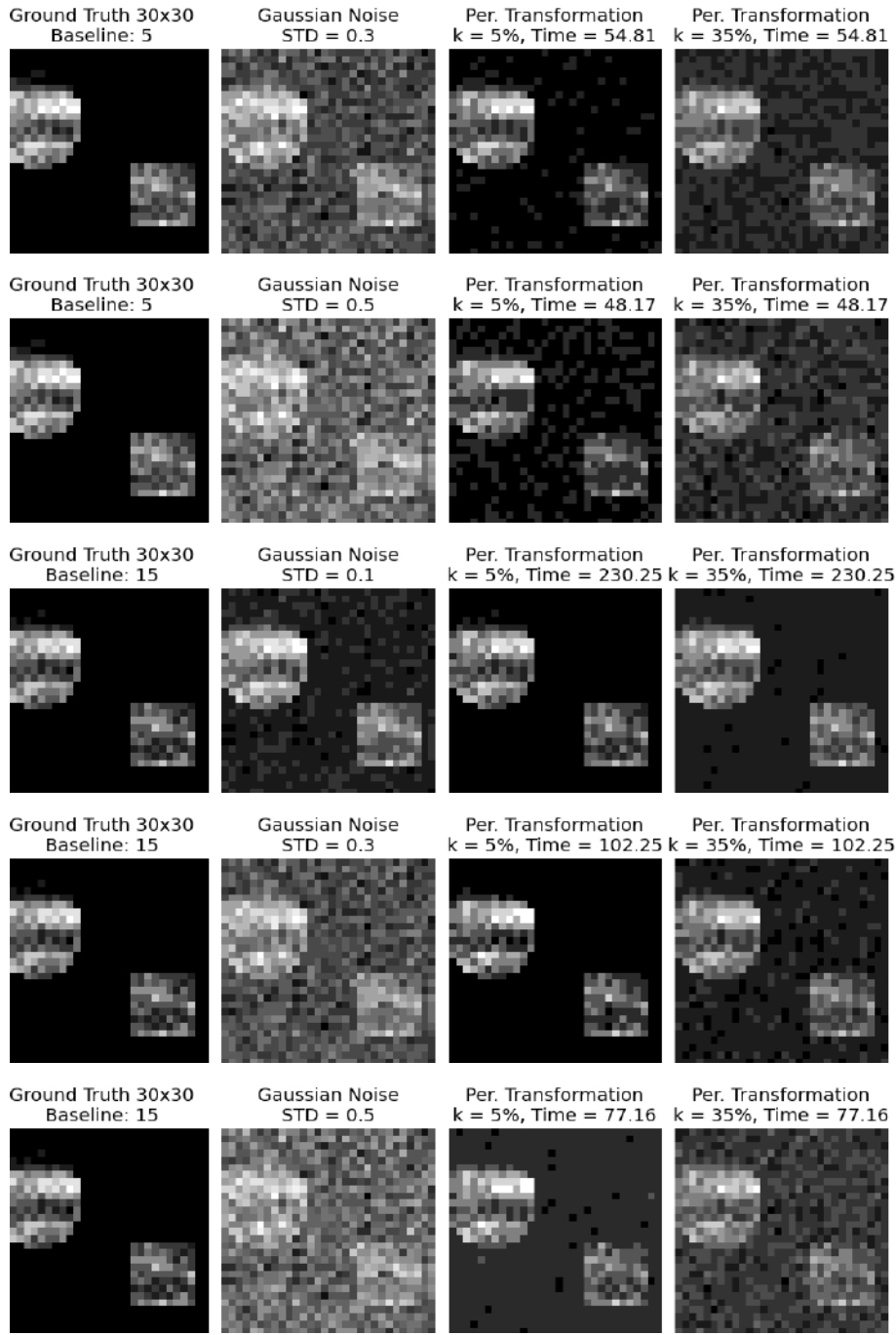

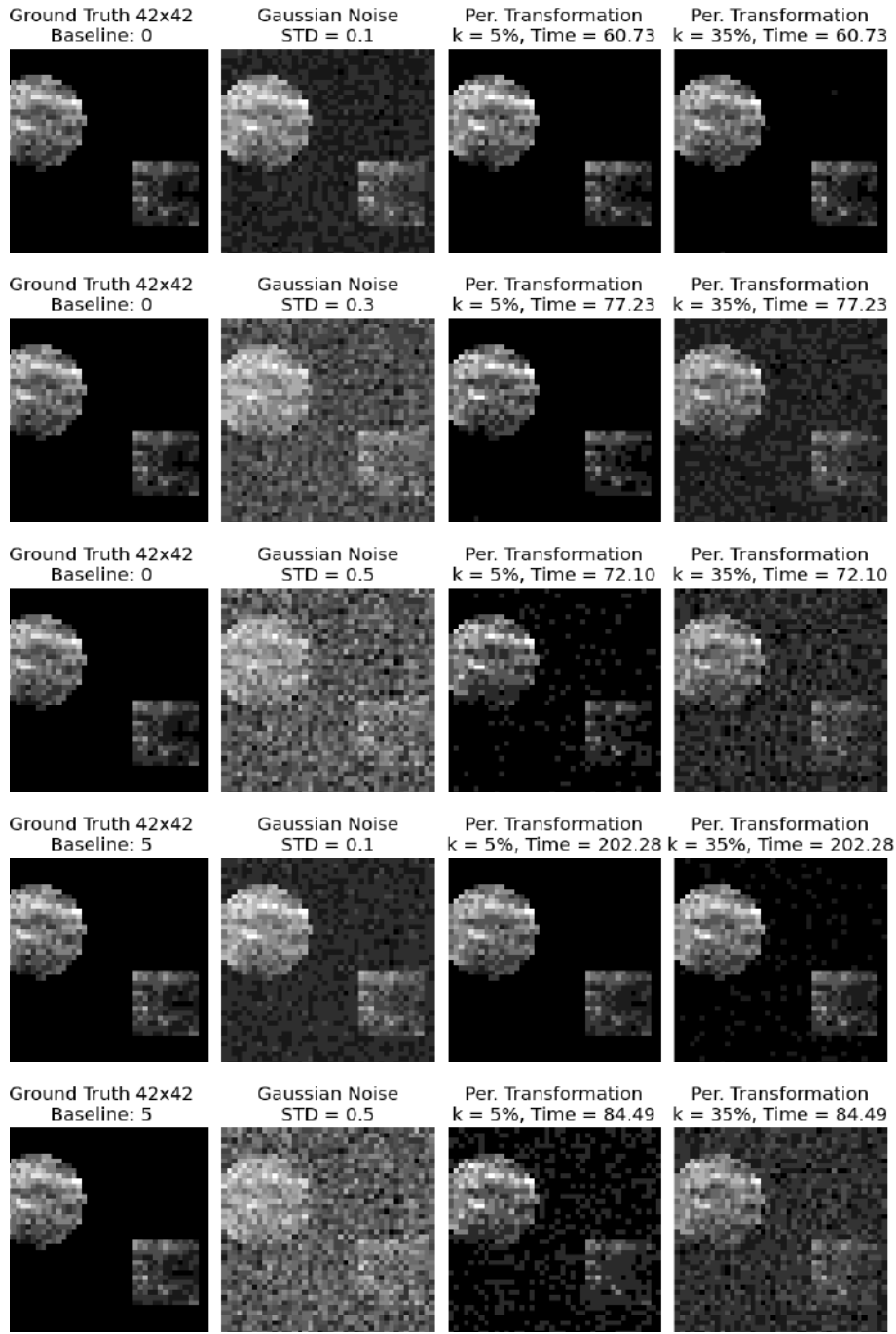

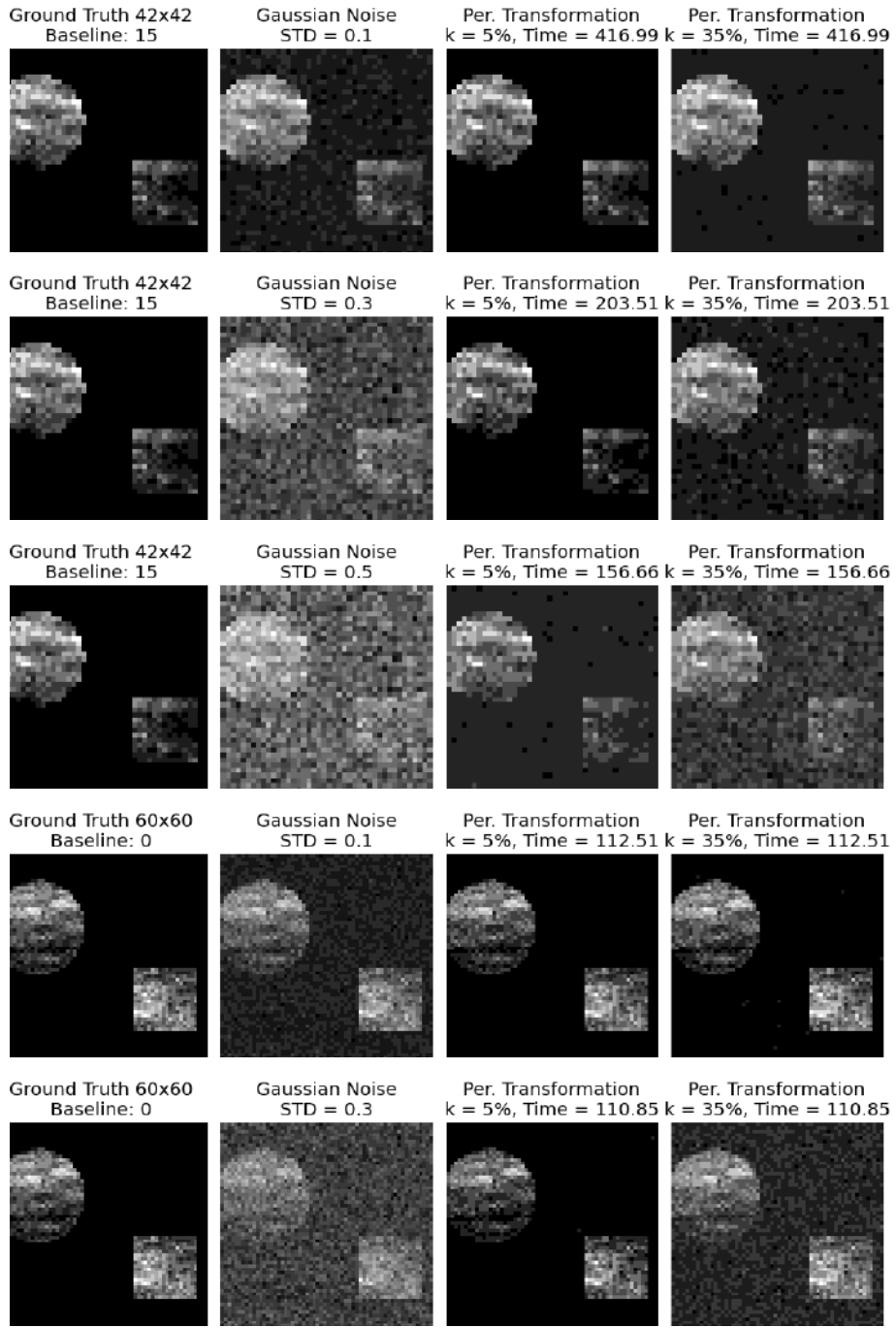
4

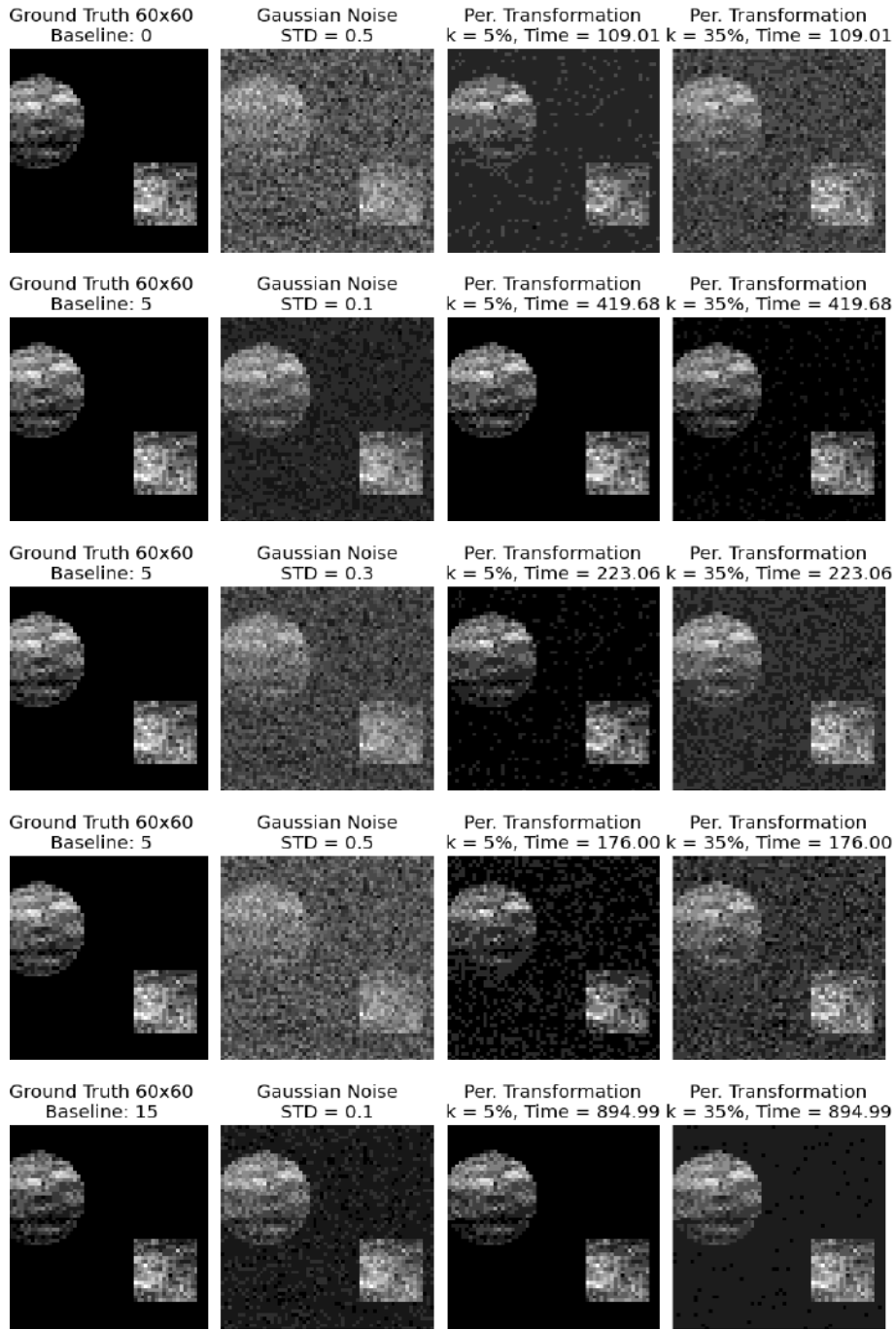

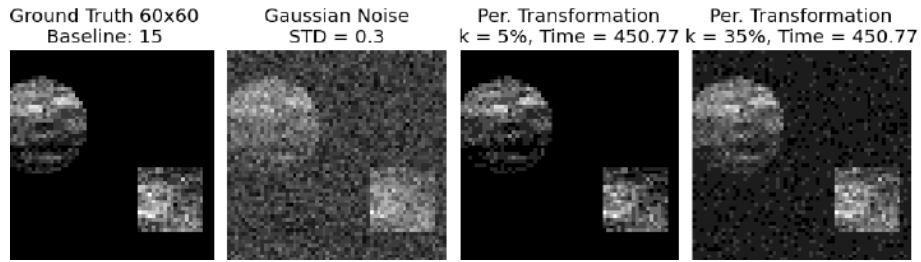

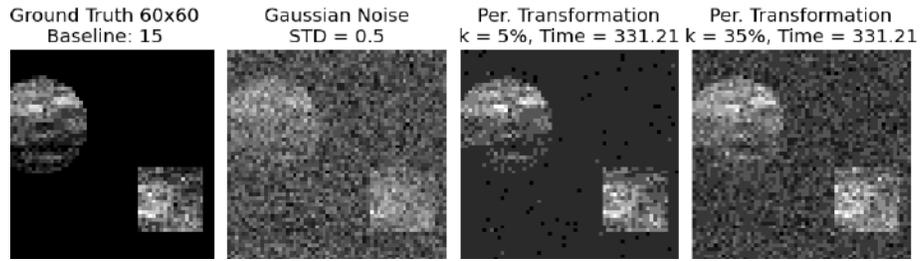

**Poisson Noise**

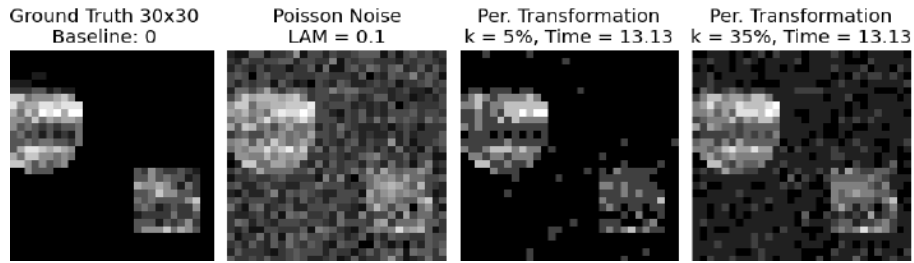

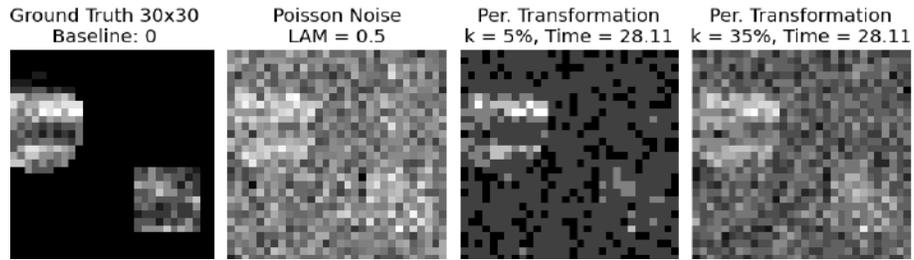

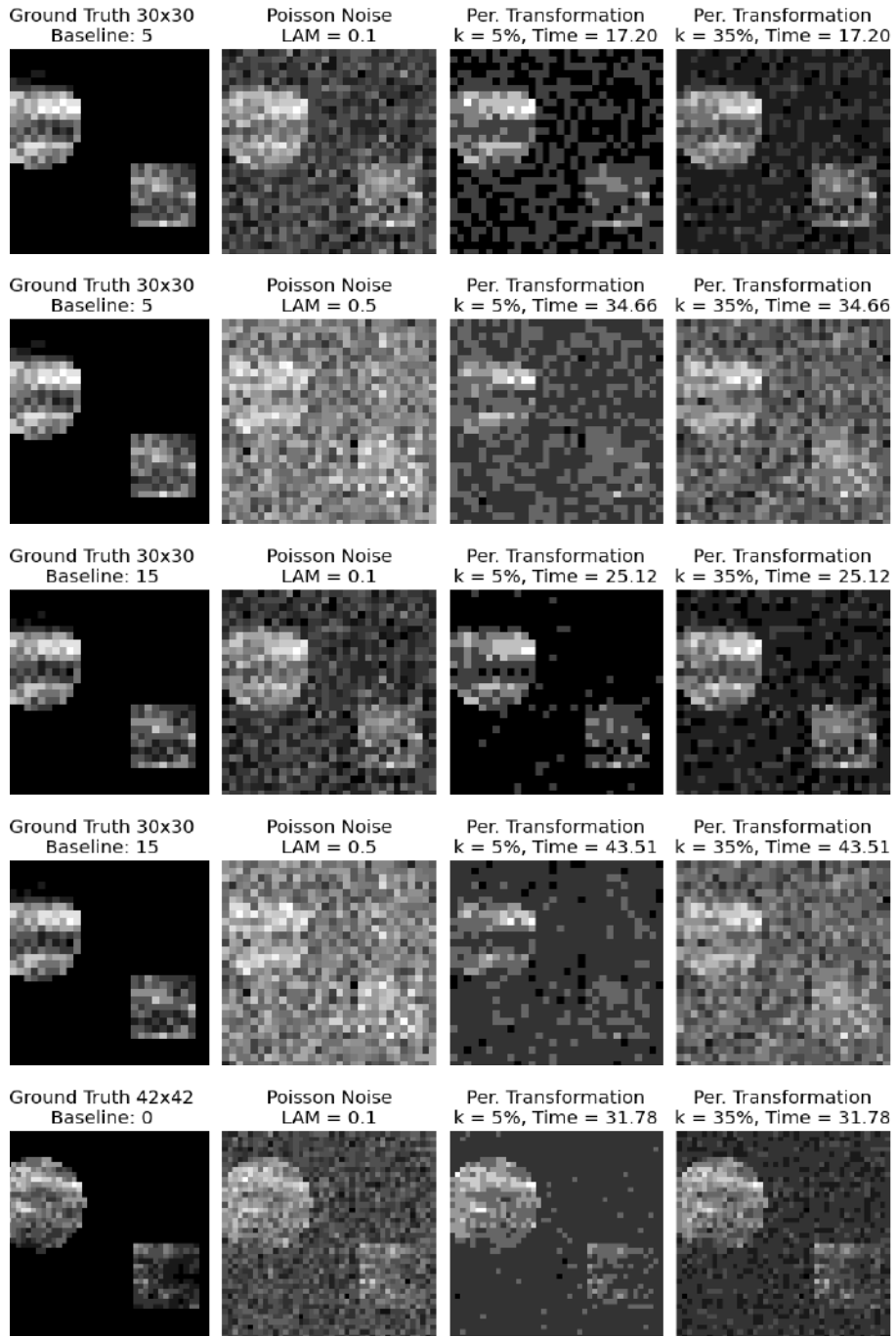

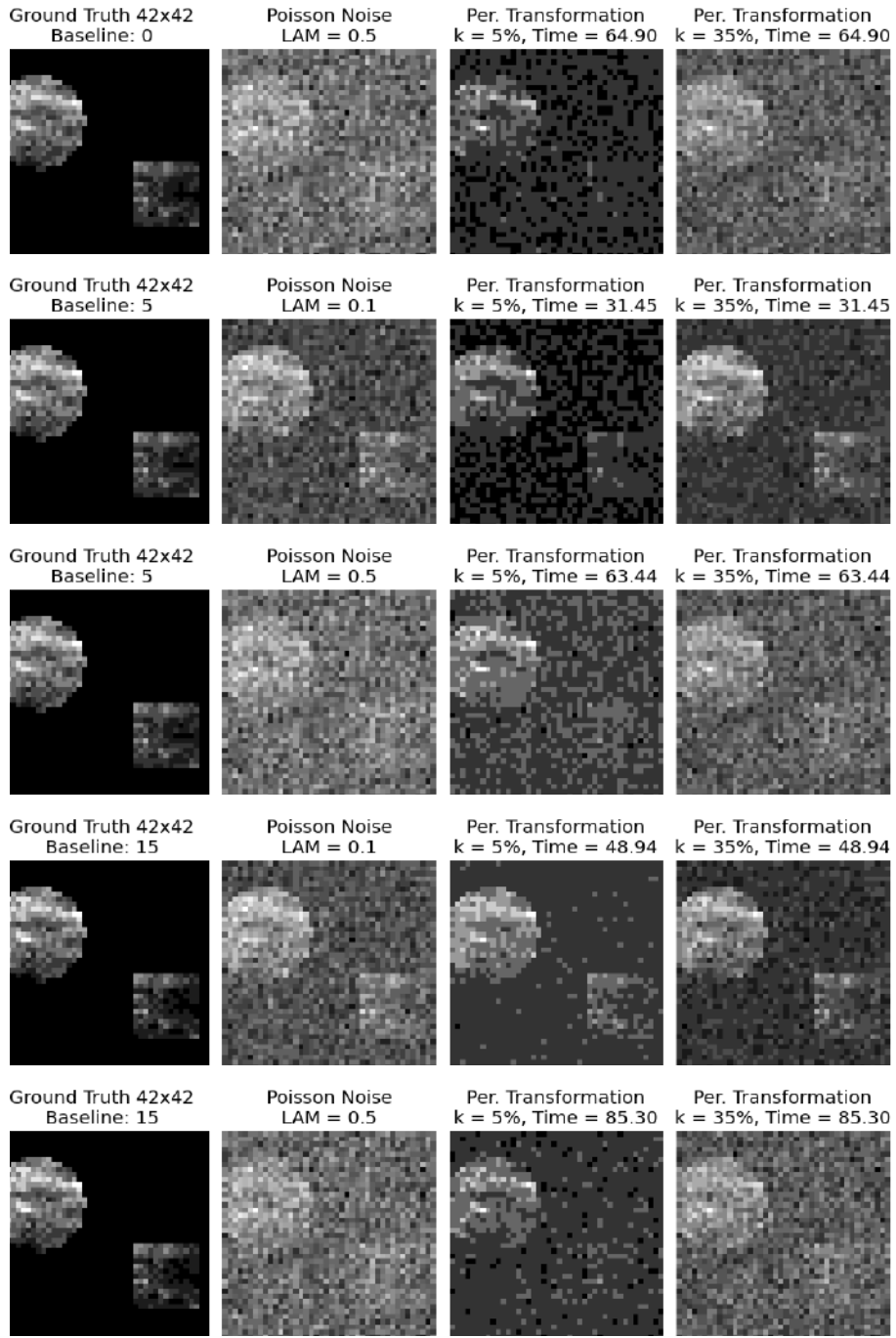

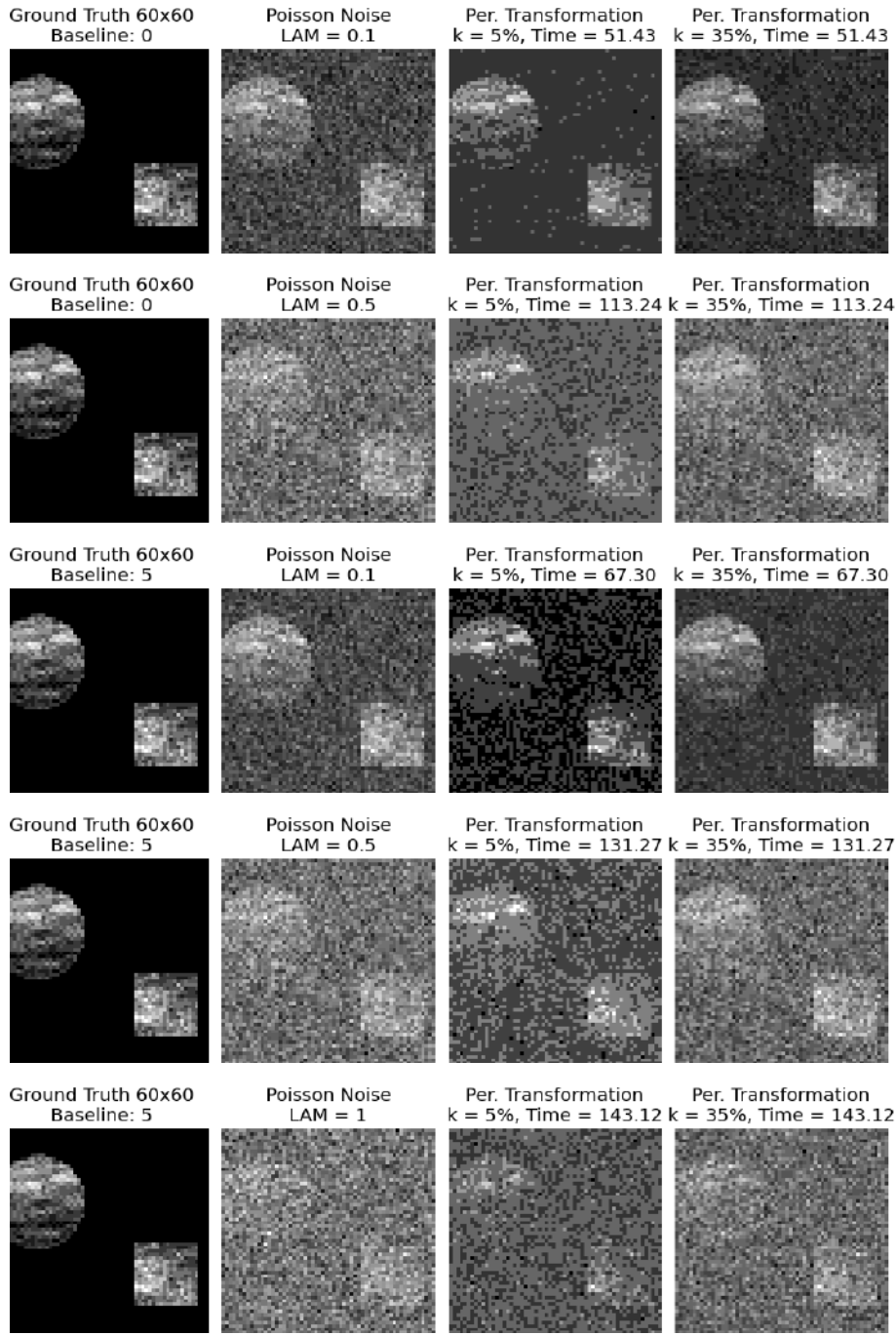